\documentclass[submission,copyright,creativecommons]{eptcs}
\providecommand{\event}{ASQAP 2025} % Name of the event you are submitting to
\usepackage[utf8]{inputenc}
\usepackage{listings}
\usepackage{cite,xspace}

\usepackage[
 svgnames,
 table,
]{xcolor}
\hypersetup{
  colorlinks = true,
  urlcolor = NavyBlue,
  citecolor = NavyBlue,
  linkcolor = NavyBlue,
 }

\xdefinecolor{clrCO}{HTML}{F15B21}
\xdefinecolor{clrCOmono}{HTML}{797979}
\xdefinecolor{clrRE}{HTML}{0F0057}
\xdefinecolor{clrREmono}{HTML}{222222}
\xdefinecolor{clrRR}{HTML}{7c7791}
\xdefinecolor{clrA}{HTML}{B0A8BA}
\xdefinecolor{clrAmono}{HTML}{A0A0A0}
\colorlet{clrCOLarge}{clrCO!90} % not tuned
\colorlet{clrRELarge}{clrRE} % tuned
\colorlet{clrALarge}{clrA!30} % tuned
\colorlet{clrTableHeaderBg}{clrRE}
\colorlet{clrTableHeaderFg}{white}
\colorlet{clrCell}{clrALarge}
\colorlet{clrEmph}{clrRE}
\colorlet{clrSectionTitle}{clrRE}
\colorlet{clrSectionCount}{clrA}
\colorlet{clrFooter}{clrRE}
\colorlet{clrFooterLine}{clrA}
\colorlet{clrArrayRule}{clrA}
\colorlet{clrArrayCell}{clrA!30}
\colorlet{clrBoxBg}{clrALarge}
\colorlet{clrBoxFg}{black}
\colorlet{clrCaption}{clrRE}
\colorlet{clrBox}{clrRE}
\colorlet{mygray}{clrRE!75}
\definecolor{OliveGreen}{RGB}{0,110,51}

\usepackage{iftex}
\usepackage{graphicx}
\usepackage{tikz}
\usetikzlibrary{shapes.geometric, arrows, fit, positioning}

\tikzstyle{data} = [rectangle, minimum width=2.5cm, minimum height=1cm, text centered, draw=black]
\tikzstyle{api} = [draw=black, thick, fill=gray!10, rounded corners, inner sep=0.5cm]
\tikzstyle{arrow} = [thick,->,>=stealth]

\usepackage{amssymb}
\usepackage{amsmath}
\usepackage{mathtools}
\usepackage[tone]{tipa} % phonetic IPA characters
\newcommand{\R}{\mathbb{R}}

\usepackage[draft,silent]{fixme} % add comments in the margin
\fxsetup{theme=color,mode=multiuser}
\definecolor{fxtarget}{rgb}{0.8000,0.0000,0.0000}
\FXRegisterAuthor{pk}{PK}{Paul}
\FXRegisterAuthor{rs}{RS}{Riccardo}
\FXRegisterAuthor{lt}{LT}{Liz}

\usepackage[binary-units]{siunitx}

\ifpdf
  \usepackage{underscore}         % Only needed if you use pdflatex.
  \usepackage[T1]{fontenc}        % Recommended with pdflatex
\else
  \usepackage{breakurl}           % Not needed if you use pdflatex only.
\fi

\newcommand{\etal}{\emph{et al.}\xspace}
 \def\myttsize{\fontsize{8}{8}}
 
\lstdefinelanguage{ABS}{keywords={for, if, else, while, this,return,get,new,class,interface,implements,if,await, skip,suspend}, sensitive=true, comment=[l]{//}, morecomment=[s]{/*}{*/}, morestring=[b]"}

\lstdefinestyle{absstyle}{
language=ABS,columns=fullflexible,
 		   mathescape=true,%
 		   showstringspaces=false,%
keywordstyle=\bf\sffamily\color{NavyBlue},
commentstyle=\sl\sffamily,%
basicstyle=\small\sffamily,
inputencoding=latin1, % i would prefer utf8
extendedchars,xleftmargin=2em
}

\newenvironment{absinline}[1][]{%
\lstset{language=ABS,
		 showstringspaces=false,
		 columns=fullflexible,mathescape=true,%
         keywordstyle=\bf\sffamily,commentstyle=\sl\sffamily,%
         basicstyle=\normalsize\sffamily,inputencoding=latin1, % i would prefer utf8
         extendedchars,xleftmargin=2em,#1}}
{}

\lstnewenvironment{absexample}{
~\\
\noindent 
\lstset{style=absstyle,
basicstyle=\myttsize\sffamily,
firstnumber=last,
framerule=0pt,
frame=tblr,
xleftmargin=4pt,
}}
{}

\lstnewenvironment{absexamplen}[1][]{
\lstset{style=absstyle,
basicstyle=\myttsize\sffamily,
framerule=1pt,tabsize=2,
backgroundcolor=\color{Green!5},
rulecolor=\color{Green!80!black},
frame=tblr,
captionpos=b,
xleftmargin=4pt,#1
}}
{}

\title{BedreFlyt: Improving Patient Flows through\\ Hospital Wards with Digital Twins}
\author{Riccardo Sieve
\institute{University of Oslo\\
Oslo, Norway}
\email{riccasi@ifi.uio.no}
\and
Paul Kobialka
\institute{University of Oslo\\
Oslo, Norway}
\email{paulkob@ifi.uio.no}
\and
Laura Slaughter
\institute{University of Oslo\\
Oslo, Norway}
\email{l.a.slaughter@dscience.uio.no}
\and
Rudolf Schlatte
\institute{University of Oslo\\
Oslo, Norway}
\email{rudi@ifi.uio.no}
\and
Einar Broch Johnsen
\institute{University of Oslo\\
Oslo, Norway}
\email{einarj@ifi.uio.no}
\and
Silvia Lizeth Tapia Tarifa
\institute{University of Oslo\\
Oslo, Norway}
\email{sltarifa@ifi.uio.no}
}
\def\titlerunning{BedreFlyt}
\def\authorrunning{R.~Sieve et al.}
\begin{document}
\maketitle

\begin{abstract}
  Digital twins are emerging as a valuable tool for short-term
  decision-making as well as for long-term strategic planning across
  numerous domains, including process industry, energy, space,
  transport, and healthcare. This paper reports on our ongoing work on
  designing a digital twin to enhance resource planning, e.g., for the
  in-patient ward needs in hospitals.  By leveraging executable formal
  models for system exploration, ontologies for knowledge
  representation and an SMT solver for constraint satisfiability, our
  approach aims to explore hypothetical ``what-if'' scenarios to
  improve strategic planning processes, as well as to solve concrete,
  short-term decision-making tasks. Our proposed solution uses the
  executable formal model to turn a stream of arriving patients, that
  need to be hospitalized, into a stream of optimization problems,
  e.g., capturing daily inpatient ward needs, that can be solved by
  SMT techniques.  The knowledge base, which formalizes domain
  knowledge, is used to model the needed configuration in the digital
  twin, allowing the twin to support both short-term decision-making
  and long-term strategic planning by generating scenarios spanning
  average-case as well as worst-case resource needs, depending on the
  expected treatment of patients, as well as ranging over variations
  in available resources, e.g., bed distribution in different
  rooms. We illustrate our digital twin architecture by considering
  the problem of bed bay allocation in a hospital ward.
\end{abstract}

\section{Introduction}\label{sec:intro}

Digital Twins (DTs) are virtual information constructs that capture
the structure, context, and behavior of the system they are twinning,
are dynamically updated with data from the twinned system, have
predictive capability, and inform decisions that realize value,
according to a recent definition by the National Academy for Science,
Engineering and Medicine (NASEM) ~\cite{nasem24}.  While many
applications of DTs so far can be found in engineering disciplines,
based on the idea of creating an increasingly accurate ``virtual
replica'' of a physical system to predict behavior by means of
sophisticated simulation techniques and a closed feedback loop to a
twinned cyber-physical system (e.g., \cite{fitzgerald24edt}), this
recent definition is broader.  In fact, DTs are today used in a
variety of domains outside of cyber-physical systems, including
healthcare~\cite{vallee-digital-2023},
manufacturing~\cite{DBLP:journals/rcim/BilleyW24}, and
transportation~\cite{DBLP:journals/tits/ChangZMF24}.  We believe DTs
have significant potential as a tool for the model-driven exploration
of so-called ``what if'' scenarios, moving from the predictive
analysis of near-future events to the prescriptive analysis of
hypothetical scenarios. Thus, DT technology appears to be useful both
for short-term decision-making and long-term strategic planning in
various domains.\looseness=-1

If we consider DTs from a formal methods perspective, DTs differ from
standard model-driven techniques by supporting the dynamic update of
the model, leveraging live data from the modeled system (often
referred to as the physical twin). We may think of the DT as an
infrastructure for data-driven formal methods (e.g.,
\cite{kobialka24fm}), in which the stream of data from the twinned
system is used to configure a formal model. Similarly, the what-if
scenario requested by the user of the DT may determine other aspects
of the model as well as the properties to be analyzed. This way, the
DT infrastructure may be thought of as a self-adaptive
system~\cite{self-adapt} for advanced model management, generating the
models and determining the analyses to be performed over these models.

Our focus here is on DTs for resource management in healthcare, a
domain in which resource management is crucial to efficient operations
\cite{hospitals-competition}.  The proper handling of resources at a
hospital is concerned with how, e.g., trained staff, bed availability
in the ward, and necessary rooms and equipment match the needs of the
different activities of the hospital, such as the treatment of
patients.  Efficient resource allocation provides a structured way to
better manage workflow and adjust it dynamically, avoiding bottlenecks
in operations, thereby allowing for better prioritization and
utilization of staff~\cite{resource-hospitals}.  Simulations have been
successfully used to improve resource allocation in a hospital
\cite{simulation-resources}; a DT can connect such simulation models
to live data to ensure that the simulation models do not deviate from
the resource allocation problem of the hospital.  This way, a DT can
become the point of contact between static planning and dynamic
optimization, allowing a better management of the workflow and making
it possible to adjust it dynamically. By configuring the models to
explore different scenarios, it is possible to compare different
strategies under different assumptions with respect to resources as
well as incoming patients.

This paper discusses the initial design of \emph{BedreFlyt}
(\textipa{\sffamily /"be:dr@ fly:t/}), a DT for resource planning in a
hospital ward, with a particular focus on bed bay allocation. The DT
combines a knowledge base formalizing knowledge about patient
treatments and the ward, an actor-based executable formal model to
explore scenarios for streams of incoming patients with associated
treatments, and a constraint solver to perform the actual bed
allocation.  Technically, we combine the executable modeling language
ABS~\cite{DBLP:conf/fmco/JohnsenHSSS10,johnsen15jlamp} with the SMT
solver Z3~\cite{de2008z3} and knowledge graphs~\cite{ji22tnnls}. The
orchestration language
SMOL~\cite{DBLP:conf/isola/KamburjanKSTCJ22,kamburjan21eswc} is used
to combine knowledge graphs with the ABS and Z3 models. SMOL supports
querying a knowledge base, which includes the reflection of the
runtime state of the SMOL program itself, via SPARQL and SHACL
queries. The ABS model transforms the stream of patient data into a
stream of constraint problems that capture the bed allocation problem
at different points in time. Combined with a description of the ward,
these are turned into SMT problems, which we give to Z3. The result is
the multi-day planning of bed allocation for patients, that minimizes
bed reallocation. We evaluate the design on a realistic patient
treatment scenario, based on a historical dataset for a hospital ward
at
Rikshospitalet,\footnote{\url{https://www.oslo-universitetssykehus.no/steder/rikshospitalet/}}
and further consider how the design scales to larger scenarios up to
2000 patients.

\emph{Main contributions:} a DT design combining formalized knowledge,
executable formal models and SMT
to explore ``what if''
scenarios, its implementation, and its application to a hospital ward
scenario.\looseness=-1

\emph{Paper overview:} Section~\ref{sec:ex} motivates our work by a
hospital ward planning problem, Sect.~\ref{sec:bg} reviews background
and Sect.~\ref{sec:approach} gives an overview of our DT design.  Then
Sect.~\ref{sec:impl} discusses our prototype implementation,
Sect.~\ref{sec:exp} experimental results, and Sect.~\ref{sec:disc}
perspectives on the DT design and future work.\looseness=-1

\section{Motivating Scenario}\label{sec:ex}

Resource planning in hospitals is surprisingly complex. Already the
problem of allocating beds to patients who are admitted for multi-day
stays at a hospital ward is a dynamic scheduling problem concerned
with how the patients' beds are placed in so-called bed bays,
i.e. designated slots inside the different rooms of the hospital ward
where rolling beds are placed.  Additionally, the patients' bed needs
may vary over time, and are subject to numerous constraints; for
example, to allocate a bed to a bed bay in the hospital ward one needs
to consider whether the patient requires continuous monitoring from
staff and the extent to which the patient can share a room with other
patients.

Today, when patients are admitted for a multi-day stay at a hospital
ward, their admission is often handled manually.  This manual process
is still the default procedure in many major hospitals.  Not only does
this drastically increase the workload, as various requirements have
to be taken into account, it is also time- and
resource-consuming. While the nurse tries to find a good bay for the
patient's bed, other patients still require care and sometimes
immediate attention.  The admitting nurse searches for a suitable
placement for the different patients, while accounting for patients'
diagnosis, gender, and infection status, as contagious patients are to
be isolated.  The nurse also ensures that the patients' room is
appropriate given their needs; for example, more acute patients stay
closer to the watch room to facilitate monitoring.  Furthermore, a
patient's needs may vary over a multi-day stay. In fact, optimal
resource usage suggests a near-continuous reallocation of bed bays to
patients as the overall needs of the currently admitted patients
change over time. On the other hand, patients should not be moved
unnecessarily!

\begin{figure}[t]
    \centering
    \includegraphics[trim={0mm 0mm 0mm 0mm},clip,width=0.9\linewidth]{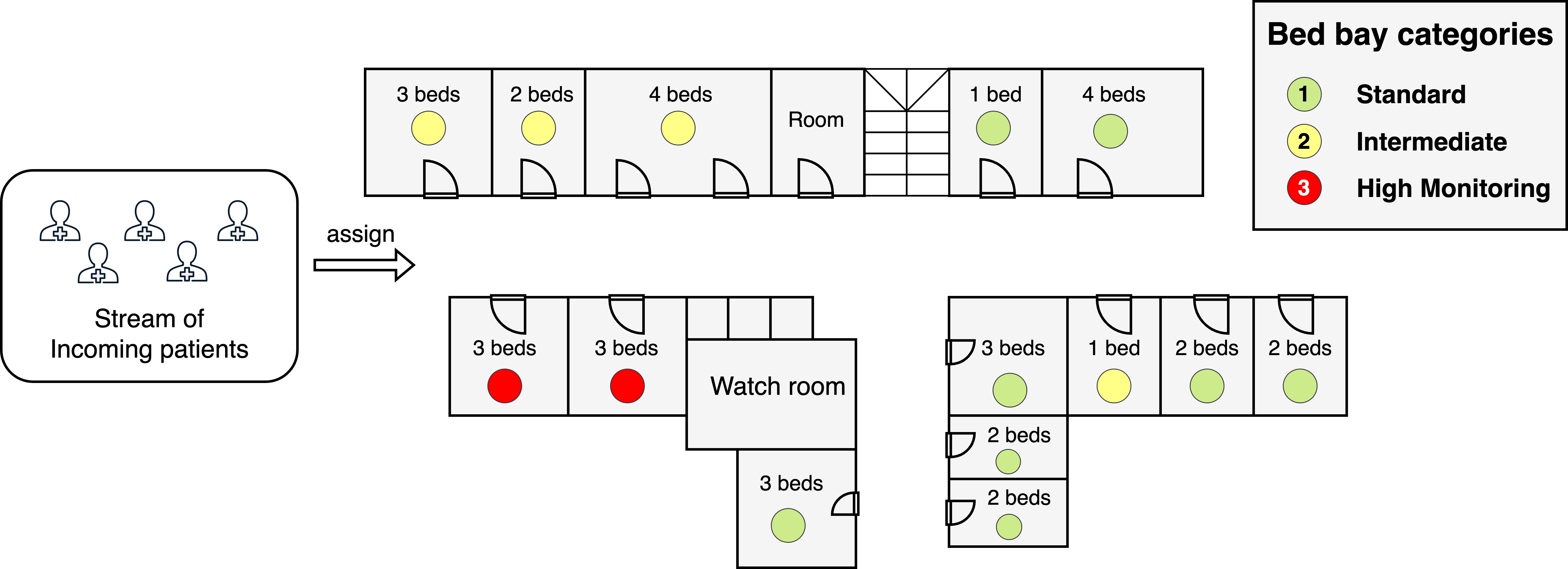}
    \vspace{-3mm}
    \caption{Example of a hospital ward.\label{fig:hospital-work}}
\end{figure}

Figure~\ref{fig:hospital-work} illustrates this bed allocation problem
for a hospital ward. A stream of patients arrive on a day-by-day basis
and get assigned a bed bay in one of the rooms of the hospital
ward. The ward is designed with rooms that satisfy different
constraints and have different capacities in terms of bed bays; in
particular, some rooms near the watch room have bed bay category
\emph{High Monitoring} and can host patients who need continuous
monitoring from staff, other rooms host patients that need
intermediate levels of attention from staff, etc.  Thus, there are
various constraints associated with the bed bays of the hospital ward,
and various needs associated with the treatments of the admitted
patients; both the constraints on the bed bays of the hospital ward
and the needs of the admitted patients can vary over
time. \looseness=-1

A DT of the hospital wards can support real-time insights into the
ward workload and support decision-making.  The DT can alleviate the
continuous process of allocating bed bays to patients, thereby
reducing the workload of staff at the hospital ward. In particular,
the twin could produce a list of patients with their room
(re-)allocations at different points in time by tracking the patients,
their treatments, and the associated bed allocation constraints.  The
number of incoming patients to the hospital ward can vary depending on
the time of day, the day of the week, and the season.  Similarly, the
twin could easily adapt to changes in the hospital ward itself, such
as the change of status of a given room from intermediate to
monitoring, rooms being temporarily out of service, etc.

Finally, the DT could not only help with day-to-day bed bay allocation
planning but also with longer-term prediction and strategic planning
based on ``what-if'' scenarios. Specifically, the hospital deals with
a combination of planned and acute patients. The treatments for acute
patients can change over time, depending on the time of day, the day
of the week, and (in Norway) the amount of ice on the streets, etc. By
exploiting domain knowledge in the creation of scenarios for
exploration, the DT can assist in longer-term planning by comparing
allocation procedures depending on seasonality.  These scenarios could
also be parameterized in risk, allowing the DT to use either
average-case or worst-case scenarios for resource needs (or even a
distribution between the two).  This way, the DT can assist in
maintaining good capacity usage in the hospital ward at acceptable
risk.  Our solution is realized in a DT architecture that combines
executable formal models with knowledge representation and constraint
solving.\looseness=-1

\section{Background and Related Work}\label{sec:bg}
We focus the discussion on analysis techniques for resource planning
and DTs in healthcare delivery.  The traditional
technique used for decision-making in healthcare is
simulation~\cite{DBLP:conf/wsc/Pitt97}, used to provide insights into
the system and help identify potential issues before they occur
~\cite{latruwe-long-term-2023}. Simulation studies such as
\cite{hajlasz-simulation-2020} are conducted to understand future load
on hospital operations, in this case, demonstrating bed availability
under forecasted demographic trends (aging population) conditions.
Garcia-Vicu\~na \etal~\cite{garcia-vicuna-estimation-2023} introduced
a flexible adaptive method for efficiently estimating the distribution
for lengths of stay and estimating bed needs from near real-time
data. They conducted simulation studies, reproducing the patient
pathway for admitting a patient to a hospital ward or the ICU. The
simulations were used to predict future bed occupancy level during a
pandemic wave.

DTs are gaining considerable traction in
healthcare~\cite{DBLP:journals/cem/AlazabKKRMBBMGA23}, especially
considering how AI can be used to develop and enhance DTs for
healthcare~\cite{DBLP:journals/widm/KaulOFJZVW23}; for instance, a
study by~\cite{DBLP:journals/access/MohamedAJK23} used DTs with
machine learning models to monitor the health of patients with lung
disease in real-time, with an accuracy of 92\%. Stenseth
\etal~\cite{stenseth23pnas} used a DT-based framework to compare the
effect of different intervention strategies on the global spread of a
pandemic, by combining hypothetical ``what-if'' scenarios with
co-simulation.  DTs have also proved to be helpful for testing
healthcare IoT applications such as medicine dispensers, while
maintaining fidelity and reliability on the system~\cite{sartaj24spe}.

The potential transformative impact of DTs on healthcare delivery has
been reviewed by Vall\a'ee~\cite{vallee-digital-2023}, who argues for
the use of DTs for resource allocation to improve operational
efficiency, including the streamlining of workflows, pinpointing
bottlenecks, and ensuring optimal utilization of resources to reduce
waiting times. Concrete examples of DTs in this domain include Mater
Private Hospital in Dublin, Ireland, which partnered with Siemens
Healthineers to create a DT of the workflow for the hospital’s
Radiology Department. Their DT’s workflow simulation demonstrated
shorter patient waiting times, faster patient turnaround, increased
equipment utilization and capacity, and lower staffing
costs~\cite{siemens-healthcare-gmbh-digital-2018}. As another example,
Oregon Health Authority (OHA) used real-time data in their Mission
Control command center to create a reliable tool to track critical
hospital resources during the COVID pandemic. GE Healthcare provided
DT tools for bed allocation and inpatient capacity monitoring across
four hospitals in Portland. Through this project, the health authority
has increased collaboration between the hospitals, and they are now
able to rapidly react to increases in hospital bed demand, with a
large impact on crisis response~\cite{merkel-statewide-2020}.
Integrating AI with DTs has also been used to improve the efficiency
in resource management, allowing for task
offloading~\cite{DBLP:journals/access/JameilA24}.

In contrast to the work discussed above, our work explores DTs as a
self-adaptive infrastructure for model management and
configuration. Our DT design combines
\emph{SMOL}~\cite{kamburjan21eswc}, a custom language for DT
orchestration, knowledge bases to formalize domain knowledge,
ABS~~\cite{DBLP:conf/fmco/JohnsenHSSS10,johnsen15jlamp} for model
simulation and exploration, and Z3~\cite{de2008z3} for constraint
solving. We briefly introduce these components below.

\emph{SMOL} is an imperative domain-specific language that leverages
ontologies~\cite{DBLP:journals/expert/ShadboltBH06}
to develop DTs that realize semantic
reflection~\cite{kamburjan21eswc}: the runtime state of a \emph{SMOL}
program can be automatically represented as a knowledge graph,
and
queried from within the program itself. This technique allows
the runtime state of the DT orchestrator to be combined with domain
knowledge to, e.g., configure simulation scenarios. The domain
knowledge needed for  model management is formalized in a knowledge
base expressed in the Resource Description
Framework\footnote{\url{https://www.w3.org/RDF}} (RDF) and RDF Schema
(RDFS) for graph representation of the data, and the Web Ontology
Language\footnote{\url{https://www.w3.org/OWL/}}
(OWL) for its expressiveness alongside Description
Logic~\cite{DBLP:conf/dlog/2003handbook}. \emph{SMOL} leverages domain
expertise formalized in the knowledge base to configure (and
re-configure) models for different analyses
\cite{DBLP:conf/isola/KamburjanKSTCJ22}, effectively turning the DT
into a self-adaptive system \cite{self-adapt}. In previous work,
self-adaptation in \emph{SMOL} has been used to, e.g., autonomously
tackle lifecycle management in DTs
\cite{DBLP:conf/seams/KamburjanSBABOJ24,kamburjan24edt}.

\emph{ABS} is used to add simulation capabilities to our DT.
ABS~\cite{DBLP:conf/fmco/JohnsenHSSS10} is a timed, resource-aware,
actor-based modeling language with a formal semantics
\cite{johnsen15jlamp}, that combines a functional layer for
computation with an imperative layer for communication and
synchronization between cooperatively scheduled processes.  ABS has
been used to model, e.g., resource-aware computational workflows on
virtualized cloud infrastructure~\cite{albert14soca,lin16fase},
railroad infrastructure~\cite{kamburjan18scp} and user
journeys~\cite{kobialka24aol}. In ABS, we can model simulations
through an actor-based approach with explicit suspension points for
scheduling~\cite{DBLP:journals/scp/SchlatteJKT22}.

\emph{Satisfiability Modulo Theories} (SMT) is used to add
optimization capabilities to our DT. SMT solving provides decision
procedures for deciding the satisfiability of a formula over a theory,
e.g. linear real
arithmetic~\cite{abraham2017smt,kroening2016decision}.  If the formula
is satisfiable, the SMT solver returns a satisfying assignment for all
variables, if the formula is unsatisfiable, an unsatisfiable core can
be computed.  In our work, we use the state-of-the-art SMT solver
Z3~\cite{de2008z3} for deciding satisfiability of formulas over
quantifier-free linear real arithmetic, i.e. formulas of the form
$\exists x, y \in \R : (x + y \leq 2) \land (2\cdot x + y \geq 3 )
\land (y = 0 \lor y = 1)$.  In this case, Z3 could return the
satisfying assignment $x = 2$, $y = 0$.

\section{Approach}\label{sec:approach}
\begin{figure}[t]
    \centering
\begin{minipage}{\linewidth}
  \centering
  \scalebox{1}{\begin{tikzpicture}[
   node distance=1.2mm,
   lguide/.style = {thin,clrCO},
   lflow/.style = {very thick,clrRE,<-},
   rflow/.style = {very thick,clrRE,->},
   flow/.style = {very thick,clrRE,<->},
   twin/.style = {
     inner sep=1mm,
     text width=35mm,
     font=\scriptsize\sffamily\bfseries\color{black},
     anchor=north west,
     align=center,
     draw=clrRE,
     fill=clrA!30,
     thin,
     outer sep=0mm,
     minimum height=6mm,
     % blur shadow,
     inner xsep=2mm
   },
   physical/.style = {
     inner sep=1mm,
     text width=35mm,
     font=\scriptsize\sffamily\bfseries\color{black},
     anchor=north west,
     align=center,
     draw=clrRE,
     fill=clrA!60,
     thin,
     outer sep=0mm,
     minimum height=6mm,
     inner xsep=2mm
   },
   knowledge/.style = {
     inner sep=1mm,
     text width=35mm,
     font=\scriptsize\sffamily\bfseries\color{white},
     anchor=north west,
     align=center,
     draw=clrRE,
     fill=OliveGreen!85,
     thin,
     outer sep=0mm,
     minimum height=6mm,
     % blur shadow,
     inner xsep=2mm
   },
   component/.style = {
     inner sep=1mm,
     text width=35mm,
     font=\scriptsize\sffamily\bfseries\color{black},
     anchor=north west,
     align=center,
     draw=clrRE,
     fill=clrA!60,
     thin,
     outer sep=0mm,
     minimum height=6mm,
     inner xsep=2mm
   },
   ward/.style = {
     inner sep=1mm,
     text width=35mm,
     font=\scriptsize\sffamily\bfseries\color{white},
     anchor=north west,
     align=center,
     draw=clrRE,
     fill=blue!50,
     thin,
     outer sep=0mm,
     minimum height=6mm,
     % blur shadow,
     inner xsep=2mm
   },
]

  % fixes positioning of the drawing in the image (and image's size)
  %\draw[help lines, ultra thin, draw=clrA] (-0.3,-0.3) grid (17.5,10);
  %\path[help lines] (-0.3,-0.3) grid (17.5,10);

  % \hyphenpenalty=10000

  % \draw[lguide, fill=clrCell,draw=clrA] (0,0)
  %   -| (14.4,5.5)
  %   -| cycle;

% Digital Twin
 \node[twin,xshift=-0mm,yshift=-0mm, text width=82mm,minimum height=25mm] (dt) at (0,0) {};
% Physical twin
 \node[ward,left= 20mm of dt, xshift=-0mm,yshift=-0mm, text width=15mm,minimum height=25mm] (pt) {};

 \node[component,above= -10mm of dt,xshift=0mm,yshift=0mm, text
 width=20mm] (store) {Local Data Store};

 \node[component,left= 5mm of store,xshift=0mm,yshift=0mm, text width=20mm,
 minimum height=8mm] (driver) {Simulation Driver};

 \node[component,right= 5mm of store,xshift=0mm,yshift=0mm, text
 width=20mm] (abs) {ABS Simulation Model};

 \node[knowledge,below = 5mm of store, xshift=0mm,text width=20mm, minimum height=8mm] (kb) {Knowledge Base};

 \node[component,right= 5mm of kb,xshift=0mm,yshift=0mm, text width=20mm,
 minimum height=8mm] (solver) {Z3 Constraint Solver};

 \node[component,left= 5mm of kb,xshift=0mm,yshift=0mm, text
 width=20mm] (smol) {SMOL Application};

 \node[above= 0mm of pt,xshift=15pt] (text3)
 {\begin{minipage}{0.18\textwidth}\scriptsize\sffamily\bfseries
     Hospital Ward\end{minipage}};

 \node[above= 0mm of dt,xshift= -28mm] (text3)
 {\begin{minipage}{0.18\textwidth}\scriptsize\sffamily\bfseries
     BedreFlyt DT\end{minipage}};

 \draw[rflow,blue!60] ([yshift=8pt]pt.east) -- ([yshift=8pt]dt.west) node[midway,above] {\begin{minipage}{0.06\textwidth}\scriptsize\sffamily\bfseries Patient Stream\end{minipage}};; 
 \draw[lflow,blue!60] ([yshift=-8pt]pt.east) -- ([yshift=-8pt]dt.west) node[midway,below] {\begin{minipage}{0.084\textwidth}\scriptsize\sffamily\bfseries Bed\,\,Bay Allocation Stream\end{minipage}};; 
\draw[lflow] (smol.east) -- (kb.west); 
\draw[rflow] (kb.east) -- (solver.west); 
\draw[rflow] (abs.south) -- (solver.north); 
\draw[rflow] (driver.east) -- (store.west); 
\draw[rflow] (store.east) -- (abs.west); 
\draw[flow] (driver.south) -- (smol.north); 
\end{tikzpicture}}
\end{minipage}
\caption{Architecture of the DT, where the arrows indicate the flow of
  data.}
    \label{fig:DT-arch} 
\end{figure}
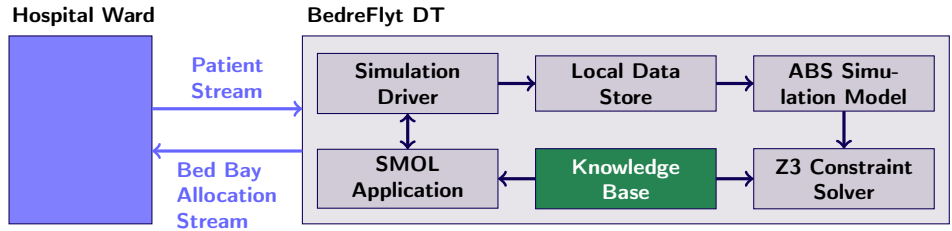

We now describe the approach we have taken to develop the DT and its
underlying components. The approach is based on the architecture shown
in Figure~\ref{fig:DT-arch}, and is driven by the following
principles:

\emph{Modularity:} The architecture is designed to be modular, and to
allow for the addition of new components and the removal of existing
ones. This allows for a more flexible and adaptable system over time,
and allows us to enforce the separation of duties.

\emph{Interoperability:} The architecture is designed for
interoperability, for integrating new components, and
for communication between existing ones. This also allows us to
enforce a separation of concerns.\looseness=-1

\emph{Scalability:} We designed the architecture with scalability in
mind, using a microservice approach.  While the performance might be
lower than a monolithic approach on a single machine, the
containerized approach allows for the deployment of the components on
multiple machines. Moreover, the use of orchestration tools to manage
the deployment makes it easier to scale the system across different
machines.\looseness=-1

\subsection{Knowledge Modeling with Ontologies}\label{sec:kb}

We use an ontology to model the assets in the twinned system, as well
as the various procedural tasks and their dependencies. An ontology is
a formal machine-readable representation of knowledge organized as a
set of concepts and the relationships between them. We build on
previous work on developing ontologies for the modeling processes and
workflows involving hospital operations~\cite{dang-ontological-2008,
  wilk-ontology-driven-2020, neumann-ontology-based-2022}.

Our ontology models bed bay capacity and the category of care-level
for each room in the ward. The ontology is used by the other
components of the system providing the standardized set of terms that
facilitates interoperability, and to feed up-to-date information about
the assets and procedural tasks to the other components. Our ontology
captures only the needed concepts and relationships for our
demonstration of the bed allocation simulator.

The ontology further models the basic structure of patient
trajectories associated with treatment flow from pre-surgery, surgery,
to post-surgery recovery. The room/bed needs through different phases
of a patient's hospital stay are related to these treatment
trajectories. The ontology design follows modeling principles to
support expansion with further knowledge about ward processes, when
this becomes needed to extend the twin's capabilities. For example,
the ontology can easily be modified to increase the range of tasks
beyond bed-related resource allocation, such as equipment needs,
imaging and diagnostics, and personnel scheduling with complex
formation of interdisciplinary teams of healthcare personnel
(see~\cite{wilk-ontology-driven-2020}).\looseness=-1

We use the ontology to capture the structure of the hospital,
specifically the ward. The ward, in the domain we used as baseline,
consists of a set of rooms $R$, each with a number of bed bays, room
$r \in R$ has $B_r$ bed bays.  Each room $r$ is associated with a bed
bay category $C_r$, used to determine the bed allocation in each room,
according to the patients' needs.  The complete set of the rooms is
associated with a specific distribution and capacity, required to
determine the availability of beds in the ward.

Moreover, the ward has a set of specific tasks that are to be
performed for a given treatment. The ontology also captures the
patients' available treatments, which are given after a patient
diagnosis. Treatments are associated with a set of tasks that need to
be performed in a predefined ordering. Each task is associated with an
average duration and a required bed bay category. For each diagnosis,
there is a corresponding patient trajectory, a deployment of a
treatment, identified by a set of tasks with their corresponding task
dependencies (that capture the ordering constraints), that need to be
performed as part of the treatment of a patient, e.g.,
\texttt{post-surgery} can happen only after \texttt{surgery}.  The
proper dependencies for the tasks are checked to ensure that the tasks
are performed in the correct order.

The ontology is then integrated into SMOL to provide the runtime
knowledge base for the DT: Each component of the asset model is mapped
to a corresponding class in the knowledge base, and the relationships
between the components are mapped to the relationships between the
classes.  The knowledge base is used to provide the data required for
the simulation model, and to provide the constraints that need to be
solved by the SMT solver. This way, no information about the ward's
resources or processes are hard-coded into the DT's simulator or
constraint-solver components; rather, these are configured on the fly
by the DT, thereby giving the DT self-adaptive capabilities \cite{self-adapt}.

\subsection{Scenario Simulation with ABS}
\label{sec:abs}

We use simulation techniques to connect the static structure of the
system (i.e., the knowledge base, see Section~\ref{sec:kb}), with
dynamic scenarios of patient flow. The workflow simulator takes a
timed stream of patients with associated treatments as input and
produces a timed stream of bed bay allocation problems as output, each
of which captures the bed bay allocation problem to be solved at a
particular point in time.\looseness=-1

The simulator, implemented in ABS, fetches from the local data
store\footnote{Technically, the local data store is realized as an
  embedded SQLite database, see \url{https://sqlite.org}.} data
produced by the Simulation Driver by combining static information from
the ontology and a desired scenario of patient inflow. The simulator
reads from the local store at different points in time and provides a
sequence of allocation problems, one for each point in time. (So far,
we inhabit the database using an anonymized, historical dataset from a
hospital ward; in the future we will investigate how to integrate live
data from such wards.)

The simulator uses the functional layer of ABS to fetch data from the
local store, including the patients and their corresponding treatment.
The simulator uses the concept of a \emph{Package} to capture ongoing
treatments. Packages consists of patient information and the remaining
tasks in the patient's treatment.  The simulation keeps track of
\emph{active} and \emph{pending} treatment tasks in a package (one per
patient). The dependencies between tasks (for example,
\texttt{post-surgery} depends on \texttt{surgery}) are also retrieved
from the knowledge base.  A task in a package becomes active once all
tasks it depends on have been performed.  While there are active
packages, the simulator performs the following steps for each point in
time:\looseness=-1
\begin{enumerate}
\item Check for new patient arrivals, and add their associated
  packages to the list of ongoing packages.
\item For each ongoing package, record information about active tasks
  (e.g., bed needs for patients, etc.).
\item For each active task in an ongoing package, decrease its
  duration or remove it from the package if its duration reaches zero
  (i.e., capturing that the task has been completed).
  \item Collect the bed needs for each patient.
\end{enumerate}
The simulation component runs as long as there are ongoing packages
and it generates a stream of bed needs per patient, for each point in
time. This output is later used to generate a stream of optimization
problems for the bed bay allocation planner.

Remark that the workflow simulator is slightly more general than our
current case study, as the architecture of the simulator can handle
tasks that occur at the same time and have different resource needs;
e.g., a laboratory test can occur while the patient occupies a bed
during recovery. Furthermore, dynamic, unforeseen variations in task
duration can be simulated by exploiting the timed semantics of ABS
\cite{johnsen15jlamp}.\looseness=-1

\subsection{SMT Solver}
\label{sec:smt}
We use Z3 to decide the satisfiability of the constraints at the
concrete ward at each point in time; i.e. for a given point in time,
in which bed bays should the beds of patients be placed such that all
constraints on gender, bed bay categories, infectious condition, and
capacity of rooms are satisfied.  For constraint solving, we
reformulate the entire problem into one quantifier-free linear real
arithmetic formula, as follows: let $R$ be the set of rooms and $P$ be
the set of patients.  For room $r \in R$, we denote with $B_r$ the
number of bed bays in the room, and with $C_r$ the category of its bed
bays.  For patient $p \in P$, we denote with $G_p$ their gender, with
$I_p$ whether or not the patient has an infectious condition that is
contagious, and with $C_p$ the patient's bed need category.  Note that
patients can only be placed in rooms with smaller or equal category,
i.e. patient $p \in P$ can only be placed in rooms $r \in R$ with
$C_p \geq C_r$, assuming an existing total ordering of bed bay
categories. We let the standard category be larger than the
intermediate category (see Figure~\ref{fig:hospital-work}), such
that a patient with a standard bed need can be placed in a room with
intermediate bed bays.\looseness=-1

To encode the constraint problem as an SMT formula, we introduce two
types of variables.  With variable $a_{pr} \in \{0,1\}$ we encode that
patient $p$ is assigned to room $r$, and with variable $g_r$ we assign
a gender to room $r$.  The assignment problem is decomposed into
several sub-formulas:
\begin{itemize}
    \item $\varphi_{patient}$ assigns each patient to exactly one room,
    \item $\varphi_{room}$ limits the number of patients in a room by the room's capacity,
    \item $\varphi_{gender}$ ensures that patients sharing a room have the same gender by enforcing that all patients in a room have the same gender as assigned to that room,
    \item $\varphi_{contagious}$ isolates contagious patients, i.e. they are alone in their room, and
    \item $\varphi_{category}$ restricts beds suitable for a patient by limiting the bed bay category.
\end{itemize}

Concluding, the formula $\varphi \coloneqq \varphi_{patient} \land \varphi_{room} \land \varphi_{gender} \land \varphi_{contagious} \land \varphi_{category}$ ensures that iff there exists an assignment from patients to bed bays, that assignment is sound.

\begin{equation*}
\begin{split}
    \varphi_{patient} \coloneqq & \bigwedge_{p \in P} \sum_{r \in R} a_{pr} = 1 ,\quad 
    \varphi_{room} \coloneqq  \bigwedge_{r \in R} \sum_{p \in P} a_{pr} \leq B_r , \quad
    \varphi_{gender} \coloneqq  \bigwedge_{p \in P, r \in R} a_{pr} \implies G_p = g_r \\
    \varphi_{contagious} \coloneqq & \bigwedge_{p \in P, r \in R} a_{pr} \land I_p \implies \bigwedge_{p' \in P \setminus \{p\}} \lnot a_{p'r} , \quad
    \varphi_{category} \coloneqq \bigwedge_{p \in P, r \in R} a_{pr} \implies C_p \geq C_r 
\end{split}
\end{equation*}

We further constrain $\varphi$ to respect the previous bed allocation
by minimizing the number of required reallocations, to avoid patients
being moved around when they have a multi-day stay at the hospital.
\looseness=-1

Let $P'$ be a new set of patients and let $v$ be a previously valid
assignment of patients to beds; i.e., $v$ is a function from
$P \cap P' \to R$ satisfying $\varphi$.  Additionally, we introduce
fresh variables $c_p$ indicating whether patient $p \in P \cap P'$ has
to be moved from room $v(p)$.  Solving
$\varphi' \coloneqq \varphi \land \varphi_{changes}$ while minimizing
the sum of $c_p$s returns a valid assignment with the least changes,
enabled in Z3 through the \emph{optimization modulo theories}
extension~\cite{bjorner2015nuz}, where
\begin{equation*}
\begin{split}
    \varphi_{changes} \coloneqq & \bigwedge_{p \in P \cap P'} a_{pv(p)} \lor c_p\ .
\end{split}
\end{equation*}

\noindent
Note that the complexity of the resulting constraint problem seems
quite high; our constraints encode a problem similar to the NP-hard
\textit{general assignment
  problem}~\cite{fisher1986multiplier}. However, our experiments in
Sect.~\ref{sec:exp} suggest that the problem is solvable in practice.

\section{Implementation}\label{sec:impl}
\begin{figure}[t]
    \centering
    \includegraphics[width=0.9\linewidth, keepaspectratio]{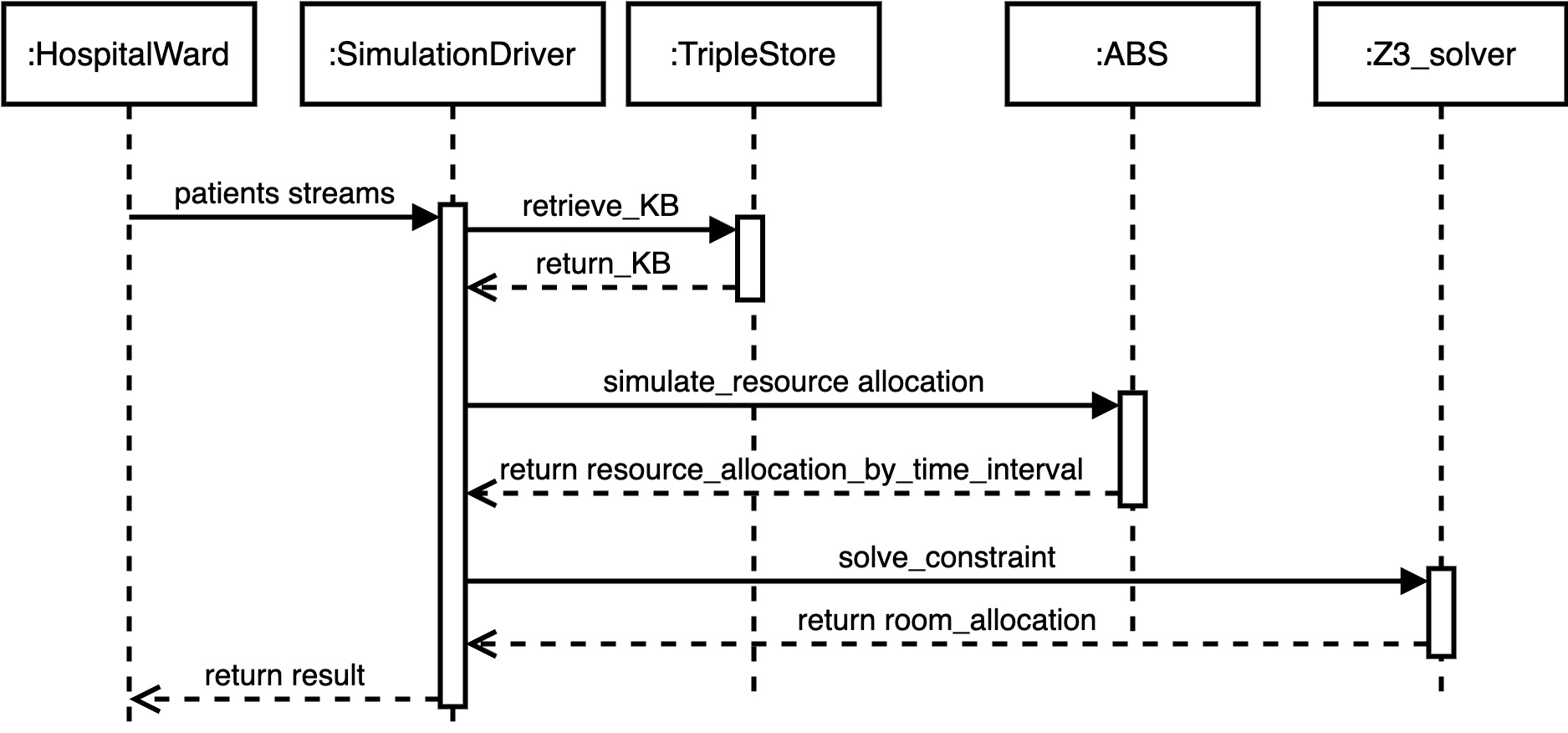}
    \vspace{-12pt}
    \caption{Sequence diagram of the bed allocation process in the
      digital twin.}
    \label{fig:seq-diagram}
\end{figure}

We now discuss the implementation of the BedreFlyt DT, which follows
the architecture depicted in Figure~\ref{fig:DT-arch}. The incoming
stream of patient data includes the patients' arrival times at the
ward, their identifier and associated treatment. The simulation
scenario handled by the \texttt{SimulationDriver} involves a multiple
steps, including data retrieval, data processing, simulation and
constraint solving, resulting in a resource allocation planner.  Each
component is developed as a separate module, with an API exposing CRUD
(Create, Read, Update, Delete) operations, and allowing interaction
with the other components.  The procedure for a single stream of
resource allocation solutions is shown in
Figure~\ref{fig:seq-diagram}.

The resource allocation planner is divided into three phases: (1) data
processing to retrieve the required data; (2) resource requirement
processing to generate the constraints; and (3) the constraint solver
process to properly define the allocation of resources to patient
trajectories.

\paragraph{Data processing phase.}

This phase consists of retrieving patient information from the patient
stream and ward information from the knowledge base created through
the ontology and processed by SMOL.  Specifically, we need the asset
model detailing the available resources of the ward, the specification
of treatments in terms of tasks with associated resource needs, the
stream of patients with their associated treatments, and (ultimately)
patient data to generate the constraint solving problems for the SMT
solver.\looseness=-1

For the resources, we focus on the rooms and their bed bays,
distinguishing the different bed bay categories, the availability of
rooms with their respective identifiers, and the bed bay capacity of
each room that is used by the constraint solver to allocate patients
to rooms.

For the tasks in a treatment and their corresponding task
dependencies, we query the knowledge base for the treatment associated
with a the patient's diagnosis, as the tasks that need to be performed
may vary. The tasks have a duration (for simplicity, we currently
consider average durations), and the resource required to perform the
task, specified by a needed bed bay category.  The ABS model is
configured with this information to collect the bed bay category needs
per patient per time interval.

The flow of patients is written to the local data store, a relational
database that can be queried through \texttt{Java Persistence API},
while information from the knowledge base is retrieved by SMOL through
SPARQL queries encapsulated into service components in the API.

The set of constraints is used as input for the constraint solver to
properly allocate the required resources to patient trajectories. In
our case, the constraints are related to the different patients that
need to be allocated to the rooms.  For each patient, we take into
consideration the severity of the diagnosis, care level requirements
(directly related to the bed bay category), and the tasks that must be
performed during each treatment, including the bed bay category
needed. Moreover, we need to check whether or not a patient may be
contagious, as this affects the patient's allocation to the different
rooms.

\paragraph{Resource requirement phase.}

The resource requirement phase is handled by the ABS model, which is
responsible for collecting the accumulated resource needs per time
interval by simulating the patient trajectories.  The Simulator
component is designed to simulate the stream of patients through
various treatment stages. The simulation starts with time interval 1,
and an empty timeline is initialized to store the bed bay needs for
each time interval. The simulation processes each package, and for
each of them, the first enabled task is executed, and the bed bay
needs are registered. If there are remaining tasks, a new package is
created and added to the remaining packages. The simulation continues
until there are no more packages to process, and the timeline for bed
bay needs is returned. The process is executed for each time interval
until all the packages have been processed.  Furthermore, an
additional step is in charge of generating the stream of constraints
per time interval. \looseness=-1

\paragraph{Constraint solver phase.}
The constraints are passed to the SMT solver as a list of
requirements, specified by the batch, i.e. the current time interval,
the patient, and the constraints for the bed bay that the patient
needs, alongside information on the rooms and their capacity.  The
simulation driver feeds information on gender, potential
contagiousness, and room numbers for the patients; the simulation
driver also provides the bed bay category of the room and the room
capacities from the knowledge base.

The constraints, expressed as a formula $\varphi_{changes}$ (see
Section~\ref{sec:smt}), are then solved by Z3, which returns the
allocation of patients to rooms if there exists a feasible allocation
solution.  The allocation is then returned to the simulation scenario
driver that forwards the output to the user of the DT.

\paragraph{Output.}

The final output consists of a stream of patient allocations to rooms
per time interval, alongside the gender of the patient. For our use
case, co-gendering in the hospital ward is against the hospital's
policy and therefore the gender requirement has been used as an
additional constraint for the room allocation.\looseness=-1

\section{Evaluation}\label{sec:exp}
We present the first proof-of-concept DT with a simulation scenario
framework that is intended to be used to plan the day-by-day bed
allocation needs for a major Norwegian hospital.  To evaluate this
first proof-of-concept, we use a stream of patient information, and
account for a sufficient ward capacity buffer for emergency patients
to produce robust plans and minimize manual labor for the allocation
of patients to rooms in a hospital ward. Working with anonymized
historical data allows us to focus the proof-of-concept on DT
functionality with a realistic flow of patient data from the hospital
ward, and defer the integration and legal challenges of actually going
online. The presented plans minimize the moving of patients while
ensuring that the needs of every patient are satisfied, if a solution
exists.  The simulation scenario driver can be used for various
time-frames, which can span up to one full year. Our DT solution
showcases extensive long-term resource planning, even computing the
bed allocation for a whole year on a day-by-day basis, under known
workloads, is computationally feasible.

In the following reported evaluation results, all scenarios are
computed on an M2 Pro MacBook Pro with \SI{32}{\giga\byte} of RAM and
Docker for the composition of the various components of the DT.

\subsection{A Realistic Hospital Scenario}
For the first experiment, we use an anonymized real-world dataset from
a hospital ward, provided by a major Norwegian healthcare provider.
In the dataset, patients are identified by their unique
\textsc{patient id}, their gender, and whether they are contagious or
not.  The diagnoses for treatments are created based on the actual
information from the real-world dataset, accounting for day-level
granularity.  Rooms are identified by their \textsc{room id}, a
synthetic identifier indicating an actual room in the hospital ward.
In the following, we call a stream of incoming patients with their
diagnosis a \emph{scenario}.

Based on the dataset, a DT of the ward is created, as described in
Section~\ref{sec:impl}.  Note that the granularity of the current
simulation is in the day-by-day needs, i.e. allocation plans are
computed per day; however, the granularity could be further reduced,
given more detailed information on the flow of patients, e.g. in which
time of the day a new patient is admitted.  All simulation scenarios
are computed based on average approximations for the task duration
perspective.  A typical scenario for the hospital ward of our use
case, with 100 patients arriving over 30 days, was computed by the DT
in approximately 30 seconds.\looseness=-1

\subsection{Synthetic Scalability of the Hospital Scenario}
\begin{figure}[t]
    \centering
\includegraphics[trim={0mm 0mm 0mm 0mm},clip,width=0.58\linewidth]{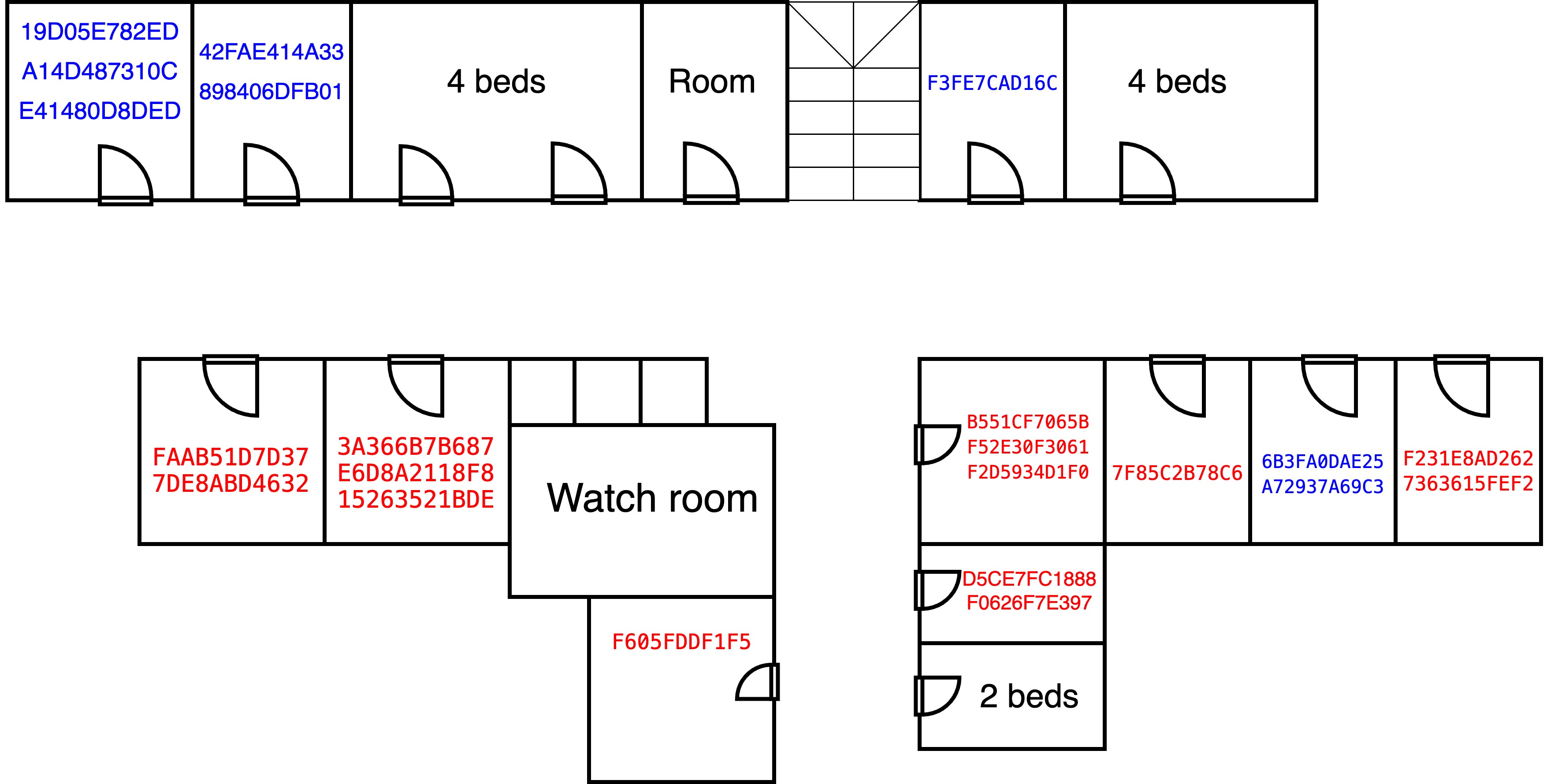}\quad
\includegraphics[trim={12mm 8mm 15mm 15mm},clip,width=0.39\linewidth]{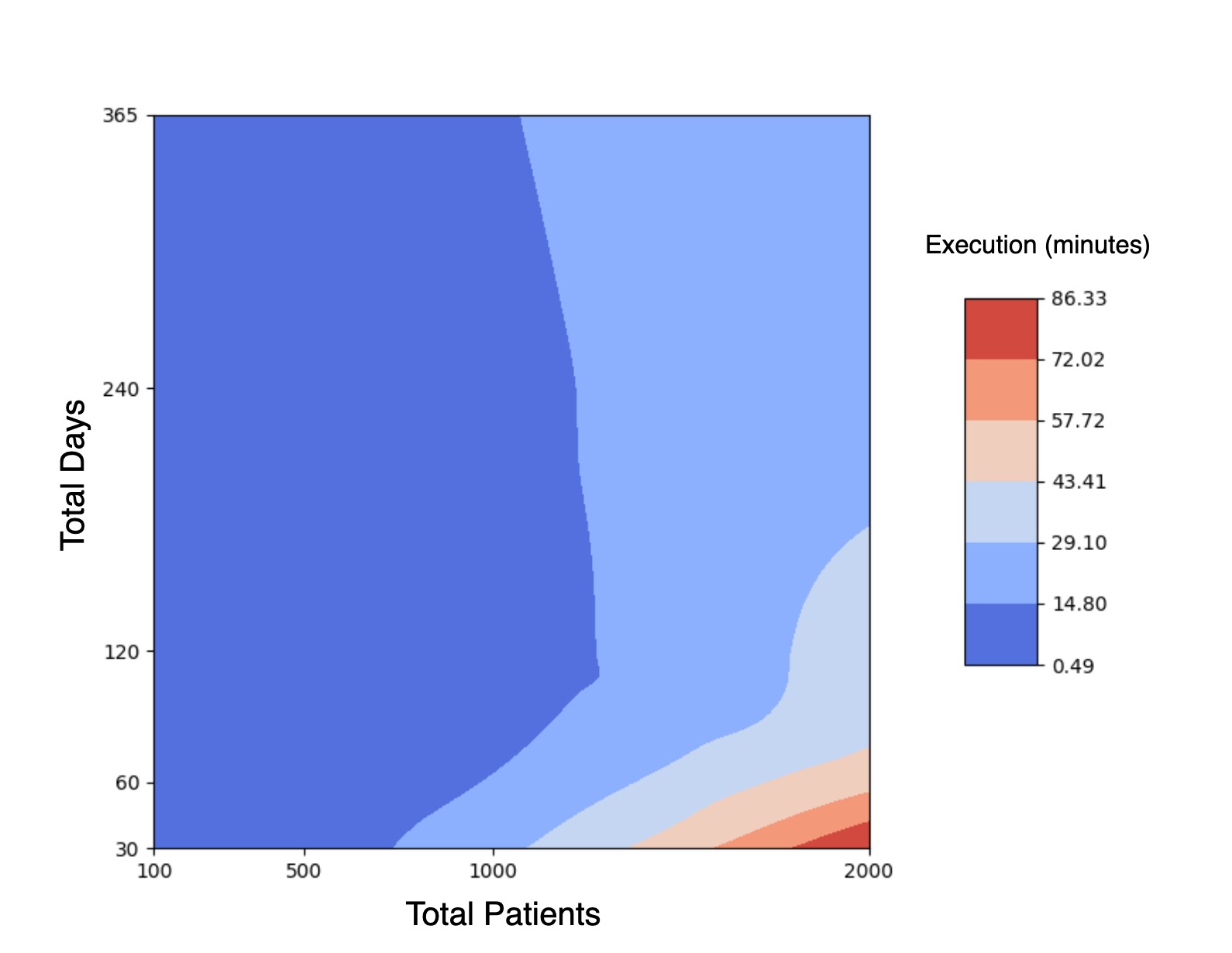}
\caption{Left: Room allocation in the ward, illustrating a solution
  from the DT for a given day (males blue, females red).  Right:
  execution times for varying scenarios ($z$-axis, in minutes),
  ranging over patients ($x$-axis, from $100$ to $2000$ patients) and
  days ($y$-axis, from $30$ to $365$
  days).\label{fig:ward-experiment}}
\end{figure}

To evaluate the scalability of our DT framework, we define multiple
scenarios by increasing the number of patients from $100$ to $2000$
and varying the number of days of the planning problem from $30$ to
$365$.  Apart from the scenario with $2000$ patients with a time frame
of $30$ days, that took $80$ minutes, the bed allocation results were
computed by the DT for all scenarios in less than $20$ minutes, as
shown by the right hand side of Figure~\ref{fig:ward-experiment}. The
reported times are the total time in minutes to compute the whole
range of scenarios, i.e., including both the simulating phase and
computing a day-by-day room allocation plan for the whole range of
scenarios within the time frame. The figure also shows a peak in
execution time for the experiments that were conducted with only a few
days but many patients. While the scenario is designed in a way that
would make it easy for the constraint solver to establish that the
problem is unsatisfiable, the high number of patients per day
increased the time needed to construct the constraint model.  In this
regard, we plan to fine-tune the model to decrease the processing
time.

In our scaling experiments, when dealing with too many patients in
high-stress test scenarios, there might not exist a feasible bed
allocation solution. When no such allocation can be provided, the day
is skipped in our experiments, and the bed allocation already in place
was left unchanged, and the simulation proceeds with the following
day.  We provide allocations for all days that are still feasible,
leaving the unsatisfied ones empty. Clearly, in a real hospital, a
more refined approach would be required here, and mitigation actions
that lie outside the scope of our DT would be needed.

\section{Discussion and Future Work}\label{sec:disc}
We have presented a proof-of-concept actor-based, simulation-oriented
digital twin framework, that leverages ontologies to create the
baseline for a knowledge base, simulation-based generation of resource
allocation problems and an SMT solver for constraints
satisfaction. The advantage of this architecture is that it is easily
configurable and its modularity easily supports further extensions.
We believe our proposed architecture combines different formal methods
in a nice way, this combination also suggests clear paths towards
interesting extensions to the work presented in this paper.

In our current work, the model of the ward is fixed, i.e. it is not
evolving over time, and models only patient trajectories within one
ward in a hospital.  We used only a small set of constraints that
could affect the resource allocation in a hospital. We here reflect on
some interesting directions for future work, based on the
proof-of-concept digital twin presented here:

\emph{Model Evolution:} Our goal is for an online twin to learn new
realistic patient trajectories over time from live data and adjust to
changes in the ward configuration. This could also mean recommending a
new ward layout that includes, for example, the addition of new bed
bays.

\emph{Constraints:} The model can easily be extended to include more
elements. For example, we might include information such as patient
age or co-morbidity. We could also consider other parameters like
equipment costs, or ease of access to resources from the hospital that
could affect particular treatments.

\emph{Granularity:} In a more realistic digital twin scenario, room
allocations will need to be made at variable rates, specified in
minutes, to obtain a more accurate representation of the hospital's
continuous operations.  We plan for the next version of the resource
allocation planner to account for real-time updates and leverage the
time semantics of ABS to support the generation of bed allocation
problems at parameterized time intervals over workflows that could be
specified with dense time.

\emph{Historic data and statistical models:} We developed the twin and
computed the simulations using a historical dataset. We have not yet
integrated statistical models into the twin architecture. The twin
model could be extended to include probabilities and distributions
that will allow more a more realistic generation of scenarios for
simulations.

An interesting extension of our work from a practical perspective, is
the actual integration of Bedre-\allowbreak{}Flyt in the workflow of a hospital
ward. We expect that connecting the twin's output to bed allocation
practice at the ward can be managed through, e.g., a visual dashboard
that allows the admitting nurse to monitor and adjust the bed
allocations suggested by the twin.  While our DT architecture has been
designed to receive a timed stream of patients with associated
treatments as input, there may be practical and legal barriers to
feeding the twin with patient data from the ward's data system that
remain to be addressed for a full online deployment of BedreFlyt.

Another interesting direction of future work, is to integrate other
resources associated with medical procedures and patient
treatment. Allocation plans would then need to not only consider the
capacity of the hospital wards but also the capacity of other
resources that the hospital needs to manage, in relation to the
patient trajectories that require hospitalization.  The next phase of
our research will consider real-time constraints, and refined
granularity of the simulation to a level of precision that also does
not inflate the complexity of the generated plans. We plan to extend
the model of the ward to include dependencies spanning over multiple
wards that influence the movement of patients. This will lead to work
on ecosystems of twins.

As discussed, these future directions raise interesting research
questions that could be tackled in a digital twin context. How do we
add complexity to the landscape of the digital twin to include other
types of user interactions with the digital twin? We envision that
considering more complex resource allocation problems, including,
e.g., healthcare personnel rostering and task scheduling, concerning
patient trajectories through the ward, will result in a digital twin
that can greatly increase efficiency in the ward.\looseness=-1

\bigskip

\noindent
\textbf{Acknowledgment.}  The work was partly funded by the
South-Eastern Norway Regional Health Authority (Helse Sør-Øst) through
the project \emph{BedreFlyt}. We thank the other project participants,
especially C\'eline Cunen, Ingrid Konstanse Ledel Solem, Frode
Strisland and Manuela Zucknik who helped collect and organize data
about patient treatments and patient streams at Oslo University
Hospital - Rikshospitalet.\looseness=-1

\bibliographystyle{eptcs}
\bibliography{references}
\end{document}